\pdfoutput=1

\documentclass[11pt]{article}
\newcommand{\best}[1]{\underline{\textbf{#1}}}  
\newcommand{\secondbest}[1]{\textit{#1}}        
\usepackage[table]{xcolor}
\definecolor{cvprblue}{RGB}{0,114,225}
\definecolor{cvprred}{RGB}{241,159,158}
\definecolor{good}{RGB}{0,150,0}  
\usepackage[most]{tcolorbox}
\usepackage{booktabs}
\usepackage{pifont}  
\usepackage{booktabs}
\usepackage{xcolor}
\usepackage[dvipsnames]{xcolor}
\usepackage{makecell}
\usepackage{booktabs}
\usepackage{siunitx}
\usepackage{amssymb}

\usepackage{multirow}
\usepackage[final]{acl}
\usepackage[ruled,vlined]{algorithm2e}
\usepackage{times}
\usepackage{latexsym}
\usepackage{amsmath}
\usepackage{graphicx}  
\usepackage[T1]{fontenc}

\usepackage[utf8]{inputenc}

\usepackage{microtype}

\usepackage{inconsolata}

\usepackage{graphicx}
\usepackage{hyperref}

\def \VersionWithComments {}
\ifdefined 
\VersionWithComments
\usepackage{marginnote}

\makeatletter
\DeclareRobustCommand\onedot{\futurelet\@let@token\@onedot}
\def\@onedot{\ifx\@let@token.\else.\null\fi\xspace}

\title{Towards Effective In-context Cross-domain Knowledge Transfer via Domain-invariant-neurons-based Retrieval}

\author{
  Jianzhi Yan\textsuperscript{1,2}\thanks{Equal Contribution},
  Zhiming Li\textsuperscript{2}\footnotemark[1],
  Le Liu\textsuperscript{1,2},
  \textbf{Zike Yuan\textsuperscript{1,2}},
  Shiwei Chen\textsuperscript{1,2},\\
  \textbf{Youcheng Pan\textsuperscript{2}},
  \textbf{Buzhou Tang\textsuperscript{2}}\thanks{Corresponding author},
  \textbf{Yang Xiang\textsuperscript{2}}\footnotemark[2],
  \textbf{Danny Dongning Sun\textsuperscript{2}}\footnotemark[2],\\
  \textsuperscript{1}Harbin Institute of Technology, Shenzhen, China \\
  \textsuperscript{2}Pengcheng Laboratory, Shenzhen, China \\
  }

\begin{document}
\maketitle
\begin{abstract}
Large language models (LLMs) have made notable progress in logical reasoning, yet still fall short of human-level performance. Current boosting strategies rely on expert-crafted in-domain demonstrations, limiting their applicability in expertise-scarce domains, such as specialized mathematical reasoning, formal logic, or legal analysis. In this work, we demonstrate the feasibility of leveraging cross-domain demonstrating examples to boost the LLMs' reasoning performance. Despite substantial domain differences, many reusable implicit logical structures are shared across domains. In order to effectively retrieve cross-domain examples for unseen domains under investigation, in this work, we further propose an effective retrieval method, called domain-invariant neurons-based retrieval (\textbf{DIN-Retrieval}). Concisely, DIN-Retrieval first summarizes a hidden representation that is universal across different domains. Then, during the inference stage, we use the DIN vector to retrieve structurally compatible cross-domain demonstrations for the in-context learning. Experimental results in multiple settings for the transfer of mathematical and logical reasoning demonstrate that our method achieves an average improvement of 1.8$\%$ over the state-of-the-art methods \footnote{Our implementation is available at \url{https://github.com/Leon221220/DIN-Retrieval}}.

\end{abstract}

\begin{figure}[!ht]
  \includegraphics[scale=0.15]{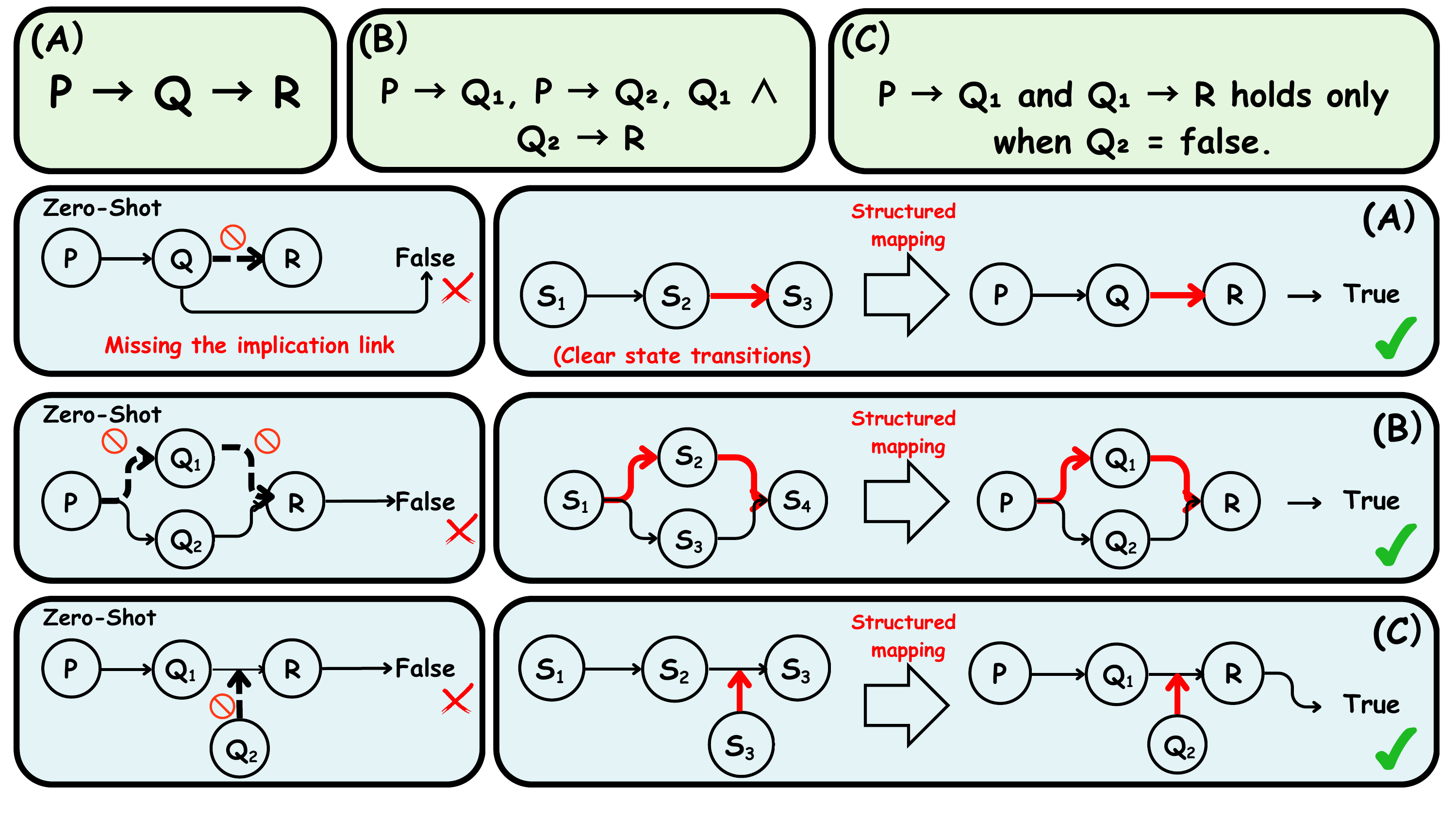}
  \caption{\textbf{Three types of failures in zero-shot LLMs.} (A) Missing intermediate links, (B) incomplete branch integration, and (C) ignored blocking conditions}
  \label{fig1}
\end{figure}

\section{Introduction}
In-context learning (\textbf{ICL}) \cite{brown2020lang,radford2019language} allows large language models (LLMs) to solve new tasks without parameter updates \cite{dong-etal-2024-survey}. With only a few demonstrations, LLMs can adapt rapidly and achieve strong performance across a wide range of tasks and domains \cite{mueller2023context,zhou2023explore,wei2022chain,lewkowycz2022solving}. While recent work has examined out-of-distribution (OOD) robustness in ICL \cite{tang2023lazy,sun2024shortcut,siska-etal-2024-examining,yuan-etal-2024-llms,honda-oka-2025-exploring,he-etal-2024-using, cheng2025revisitingreinforcementlearningllm, he2025protoreasoningprototypesfoundationgeneralizable}, these studies typically presuppose access to in-domain, expert-annotated demonstrations. Consequently, they haven't considered the practically important setting where human expertise is scarce or unavailable, and effective reasoning must instead be supported by demonstrations drawn from other domains. 

Although different domains vary in surface semantics, many reasoning tasks share underlying structural topologies \cite{Besta_2024, Besta_2025, bu-etal-2025-enhanced, zhang2024pathofthoughtsextractingfollowingpaths, li2024surveygraphmeetslarge}. Figure \ref{fig1} shows three existing types of reasoning structures that the mathematical and logical benchmarks share, yet zero-shot LLMs often fail to realize and therefore reuse them—leading to missing links, incomplete branches, or ignored blocking conditions. Notably, a cross-domain demonstration can restore the correct topology, revealing that LLMs can reuse structural reasoning patterns when appropriately guided \cite{tan2025shapereasoningtopologicalanalysis}. However, reasoning structures vary widely across tasks, making manual selection of structurally aligned demonstrations unrealistic. Thus, \textbf{improving cross-domain performance requires an automated retrieval mechanism capable of identifying examples with compatible reasoning structures.}

In this work, we conduct the first trial in demonstrating the feasibility of leveraging cross-domain examples to boost ICL performance of LLMs. To achieve effective retrieval of cross-domain samples that are of similar logical structures, we propose a novel retrieval method, called domain-invariant neurons-based retrieval (\textbf{DIN-Retrieval}) \cite{long2015learning, ganin2016domain, zhao2019learning, li2020domain, zhu2020deep}. Concretely, we identify DINs by selecting neurons whose activation polarities remain consistent across source and target domains based on cross-domain z-score statistics. These neurons define a stable DIN vector used as the retrieval representation, ensuring that similarity is computed within a domain-robust subspace. Conditioning on demonstrations selected through this invariant subspace enables more reliable cross-domain reasoning.

We validate their existence and importance via pruning: removing DINs causes significantly larger perplexity increases than random pruning, indicating their essential role in cross-domain reasoning. Building on this evidence, we then conduct cross-domain ICL experiments on mathematical (GSM8K), logical (PrOntoQA, FOLIO) transfer settings. Across all models and directions, DIN-based retrieval consistently outperforms explanation-based and embedding-based baselines, demonstrating that leveraging these invariant neurons substantially improves ICL robustness under domain shifts.

Our contributions are summarized as follows:
 \begin{itemize}
  \item We introduce DIN-retrieval, a universal neuron-level retrieval method that enables effective cross-domain in-context learning by identifying and exploiting domain-invariant neurons.
  \item Experimental results on the mutual transfer of multiple mathematical \& logical reasoning benchmarks validate that DIN-retrieval consistently enhances ICL performance.
  \item To the best of our knowledge, this work is the first to demonstrate the feasibility of using cross-domain examples for in-context learning.
\end{itemize}

We hope this work motivates future research on robust cross-domain ICL.

\begin{figure*}[t!]
  \includegraphics[scale=0.31]{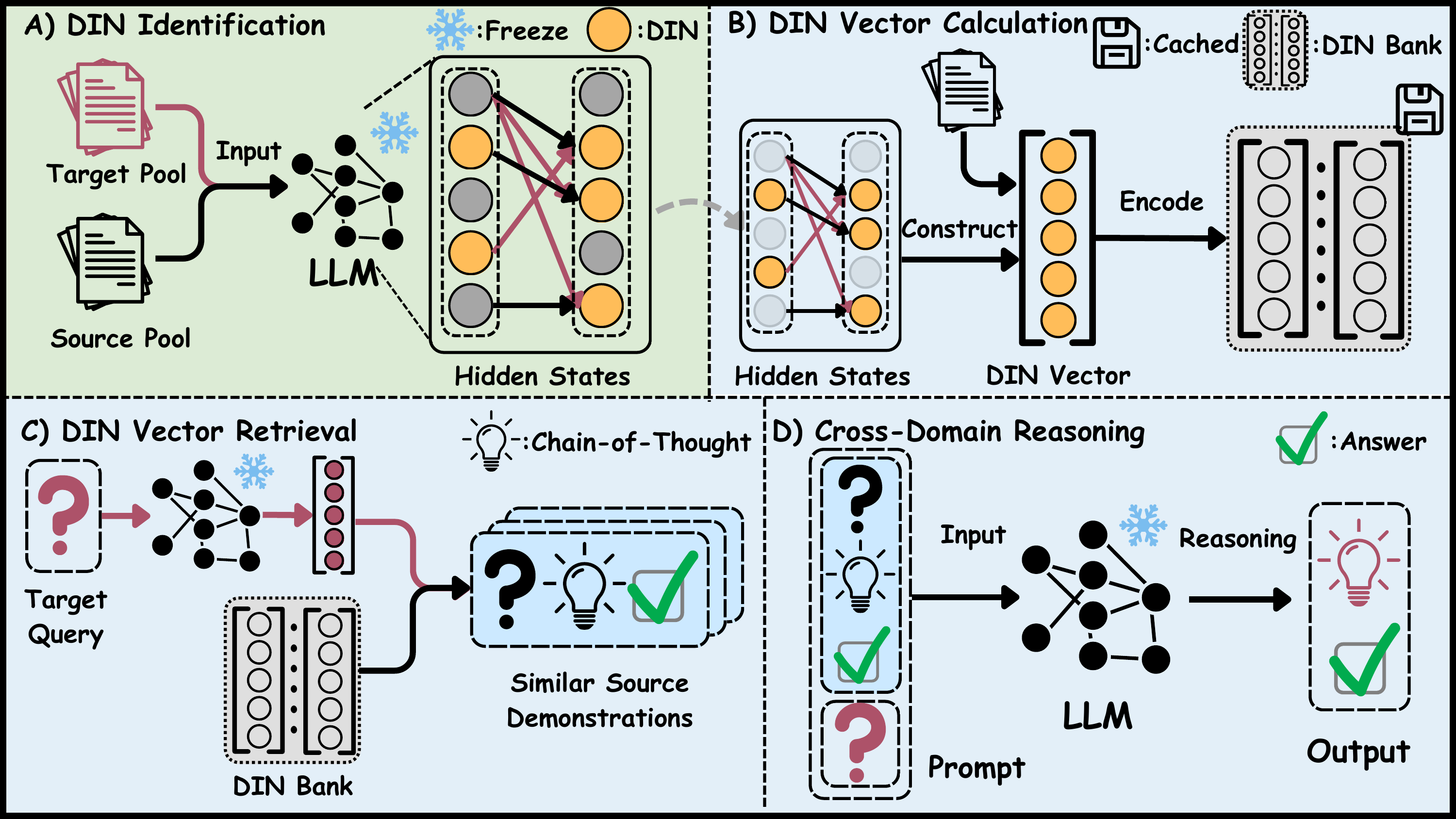}
  \caption{\textbf{Overview of the proposed DIN-based ICL framework.} The model identifies domain-invariant neurons (DINs) from source and target activations \textbf{(A)}, constructs a stable DIN vector for representation \textbf{(B)}, retrieves demonstrations via DIN vector similarity \textbf{(C)}, and performs cross-domain chain-of-thought reasoning \textbf{(D)}.}
  \label{fig:overview}
\end{figure*}

\section{Preliminaries}
In this section, we formalize the key background concepts relevant to our study: Domain Adaptation and In-Context Learning.

\subsection{Domain Adaptation}
Domain adaptation aims to transfer knowledge from a labeled source domain to an unlabeled or under-resourced target domain whose data distribution differs from the source \cite{5288526, SUN201584, farahani2020briefreviewdomainadaptation}. Let $\mathcal{D}_S = \{(x_i^S, y_i^S)\}_{i=1}^{n_S}$ denote the \textit{source domain} 
and $\mathcal{D}_T = \{(x_j^T)\}_{j=1}^{n_T}$ the \textit{target domain},
where $(x, y) \in \mathcal{X} \times \mathcal{Y}$.
Their input distributions differ:
\[
P_S(x) \neq P_T(x),
\]
while the underlying prediction function $f : \mathcal{X} \rightarrow \mathcal{Y}$ 
is assumed to be shared or related across domains.

The goal of domain adaptation is to find a mapping $f_\theta$ derived from $\mathcal{D}_S$ that generalizes to target samples $x^T \sim P_T(x)$ without access to labeled target data. A standard approach is to learn representations $h(x)\in\mathbb{R}^d$ that reduce the divergence between source and target feature distributions:
\[
\min_{\theta} \; 
\mathrm{Dist}\!\left(
    \{ h(x_i^{S}) \}_{i=1}^{n_S},
    \; 
    \{ h(x_j^{T}) \}_{j=1}^{n_T}
\right)
\]

where $\mathrm{Dist}(\cdot, \cdot)$ measures cross-domain divergence

\subsection{In-Context Learning}
In-context learning (ICL) allows large language models to infer new tasks from contextual examples \cite{wei2022chain, brown2020lang}. Unlike in-weights learning, which relies on gradient-based parameter updates, ICL adapts behavior without modifying model weights.

Formally, each training instance is linearized into an input sequence $\mathbf{x} = (x_1,\ldots,x_{|\mathbf{x}|})$ and an output sequence $\mathbf{y} = (y_1,\ldots,y_{|\mathbf{y}|})$, where each token belongs to the model vocabulary $\mathcal{V}$.  
Given a test input $\mathbf{x}_{\text{test}}$, in-context learning defines its prediction as
\[
\mathbf{y}_{\text{test}} \sim 
\mathcal{P}_{\mathcal{M}}\!\left(
\mathbf{y}_{\text{test}}
\;\middle|\;
\underbrace{
\mathbf{x}_1,\mathbf{y}_1,\ldots,\mathbf{x}_K,\mathbf{y}_K,\mathbf{x}_{\text{test}}
}_{\textbf{In-context prompt}}
\right),
\]
where the sampling operator denotes the decoding method. Each demonstration $e_i = (\mathbf{x}_i,\mathbf{y}_i)$ is drawn from a dataset
\[
\mathcal{D}=\{(\mathbf{x}_i,\mathbf{y}_i)\}_{i=1}^N.
\]

This formulation allows the model to condition on the provided examples without updating its parameters, enabling fast adaptation to new tasks without additional training cost.

\section{Method}
As aforementioned, existing ICL work ignores scenarios where human expert labelling is unavailable. In order to identify domain-invariant neurons through cross-domain alignment and use them to retrieve structurally compatible demonstrations, we present DIN-Retrieval. The following part of this section is organized as follows: we first introduce the DIN identification approach, then we illustrate the cross-domain ICL in detail.
As shown in Figure~\ref{fig:overview}, the proposed framework identifies domain-invariant neurons, constructs DIN representations, retrieves structurally aligned demonstrations, and performs cross-domain reasoning based on the retrieved examples.

\subsection{DIN Identification}\label{DIN_ident}
As shown in Part A of Figure~\ref{fig:overview}, to retrieve demonstrations aligned with the target query, we first identify neurons that are stably activated across the source and target domains. We use a \textbf{labeled source dataset} $\mathcal{D}_S={(x_i^{S}, y_i^{S})}$ and an \textbf{unlabeled target dataset} $\mathcal{D}_T={x_j^{T}}$ to identify such domain-invariant neurons. We define a class of \textbf{Domain-Invariant Neurons (DIN)} at each transformer layer. Let $h^{(l)}_t(x) \in \mathbb{R}^d$ denote the hidden state of the $t$-th token at layer $l$ for input $x$,
where $d$ is the hidden dimension and $L_x$ is the token length.
We compute the mean activation vector for a given input as:
\begin{equation}
\bar{h}^{(l)}(x) = \frac{1}{L_x} \sum_{t=1}^{L_x} h^{(l)}_t(x). 
\label{eq:mean_hidden}
\end{equation}

To measure the relative activation strength of each neuron $k$ across domains,
we compute $z$-scores using the joint statistics of both source and target samples:

\begin{align}
\mu_k &= \mathbb{E}_{x \sim (\mathcal{D}_S \cup \mathcal{D}_T)} \big[\bar{h}^{(l)}_k(x)\big], \\
\sigma_k &= \sqrt{\mathrm{Var}_{x \sim (\mathcal{D}_S \cup \mathcal{D}_T)} \big[\bar{h}^{(l)}_k(x)\big]}, \\[4pt]
z^{S}_k &= \frac{\mathbb{E}_{x\sim\mathcal{D}_S}\!\big[\bar{h}^{(l)}_k(x)\big]-\mu_k}{\sigma_k}, \label{eq:zscore-c}\\
z^{T}_k &= \frac{\mathbb{E}_{x\sim\mathcal{D}_T}\!\big[\bar{h}^{(l)}_k(x)\big]-\mu_k}{\sigma_k}. \label{eq:zscore-d}
\end{align}

Here, $z^{S}_k$ and $z^{T}_k$ quantify the standardized activation polarity of neuron $k$ in the source and target domains, respectively.

We define a set of domain-invariant neurons (DIN) at a given layer as the dimensions that exhibit consistent activation polarity across both the source and target domains, exceeding a specified z-score threshold $\tau$. Formally, the DIN candidate set is:

\begin{equation}
\begin{aligned}
\mathcal{I} =\ & \left\{k \in [1, d] \mid z_k^S > \tau \wedge z_k^T > \tau \right\} \\
& \cup\ \left\{k \in [1, d] \mid z_k^S < -\tau \wedge z_k^T < -\tau \right\}
\end{aligned}
\end{equation}

If the size of $\mathcal{I}$ exceeds a pre-defined budget $K = \lfloor k_{\text{ratio}} \cdot d \rfloor$, we select the top-$K$ dimensions with the largest combined z-score magnitude(those maximizing $|z_k^S| + |z_k^T|$).

\subsection{Cross-domain In-Context Learning}

As shown in Part B of Figure~\ref{fig:overview}, for each input $x$, we compute its DIN representation $\mathbf{v}_{\text{DIN}}(x) = \bigoplus_{l \in \mathcal{L}} h^{(l)}(x)_{\mathcal{I}^{(l)}}
$ by aggregating activations over the identified invariant neurons. Then, in Part C of Figure~\ref{fig:overview}, given a target-domain query $x_{\text{test}}^{T}$, we retrieve source-domain demonstrations based on their similarity in the Domain-Invariant Neuron (DIN) space.  The similarity between the target query and a source example is defined as:
\[
\mathrm{Sim}(\mathbf{v}_q, \mathbf{v}_i)
= \frac{\mathbf{v}_q \cdot \mathbf{v}_i}{\|\mathbf{v}_q\| \, \|\mathbf{v}_i\| + \epsilon},
\]
where $\mathbf{v}_q = \mathbf{v}_{\text{DIN}}(x_{\text{test}}^{T})$ and $\mathbf{v}_i = \mathbf{v}_{\text{DIN}}(x_i^{S})$.

To encourage diversity among selected demonstrations, we applied Maximal Marginal Relevance (MMR):
\[
\text{Score}(i)
= \lambda \cdot \cos(\mathbf{v}_q, \mathbf{v}_i)
- (1 - \lambda) \cdot \max_{j \in \mathcal{S}} \cos(\mathbf{v}_i, \mathbf{v}_j),
\]
where $\mathcal{S}$ denotes the set of already selected examples.

Finally, in Part D of Figure~\ref{fig:overview}, the top-$k$ (with $k=2$ in this work) retrieved source examples are concatenated with the target query to form the in-context prompt:
\begin{equation}
\hat{y} \sim 
\mathcal{P}_{\text{LM}}
\left(
\underbrace{[(x_1^{S}, y_1^{S}), \dots, (x_k^{S}, y_k^{S})]}_{\textbf{Source Domain}},
\ \underbrace{x_{\text{test}}}_{\textbf{Target Domain}}
\right)
\end{equation}

By retrieving demonstrations in the domain-invariant space, the resulting prompt emphasizes structurally aligned reasoning patterns, enabling more robust cross-domain generalization.

\begin{algorithm}[t]
\caption{DIN-Retrieval}
\label{alg1}
\KwIn{Source-domain pool $\boldsymbol{\mathcal{D}_S = {(x_i^S, y_i^S)}}$, Target-domain pool $\boldsymbol{\mathcal{D}_T = {x_j^T}}$,Target query instance $\boldsymbol{x_{\text{test}}}$, LLM $\boldsymbol{\mathcal{M}}$, Activation threshold $\boldsymbol{\tau}$, Neuron selection ratio $\boldsymbol{k_{\text{ratio}}}$, Number of retrieved demonstrations $\boldsymbol{k}$}
\KwOut{Model Prediction $\boldsymbol{\hat{y}}$}

\textbf{DIN Identification.}  
Compute neuron-wise statistics $(z_k^S, z_k^T)$ and select domain-invariant neurons
\begin{equation*}
\begin{aligned}
\mathcal{I} =\ & \left\{k \in [1, d] \mid z_k^S > \tau \wedge z_k^T > \tau \right\} \\
& \cup\ \left\{k \in [1, d] \mid z_k^S < -\tau \wedge z_k^T < -\tau \right\}
\end{aligned}
\end{equation*}
Construct DIN representation
$\mathbf{v}_{\text{DIN}}(x)=\bigoplus_l h^{(l)}(x)_{\mathcal{I}^{(l)}}$.

\textbf{Cross-Domain ICL.}  
Retrieve top-$k$ source examples by cosine similarity
\[
\text{Sim}(x_q,x_i)=\frac{\mathbf{v}_q\cdot\mathbf{v}_i}{\|\mathbf{v}_q\|\|\mathbf{v}_i\|},
\]
optionally refined by MMR, and predict
\[
\hat{y}\sim\mathcal{P}_{\mathcal{M}}\big([(x_1^S,y_1^S),\ldots,(x_k^S,y_k^S)],x_{\text{test}}\big).
\]

\end{algorithm}

\section{Experiments}
In this section, we evaluate the existence and usefulness of Domain-Invariant Neurons through pruning analysis and DIN-based ICL retrieval. We begin by outlining our experimental setup and then address three research questions: \textbf{RQ1} — Do DINs exist, and are they functionally important for cross-domain reasoning? \textbf{RQ2} — Is DIN-retrieval effective in improving lCL's reasoning performance? \textbf{RQ3} — How do DIN-retrieval retrieved cross-domain demonstrations boost the reasoning performance in essence?
\subsection{Experimental Setup}

\subsubsection*{Backbone Models}
We evaluated our approach using a diverse suite of open-source large language models, covering multiple architectures and scales. Specifically, we used LLaMA-3.1-8B-Instruct \cite{grattafiori2024llama3herdmodels}, Gemma-3-12B, and Gemma-3-27B \cite{gemma_2025}, as well as Qwen2.5 and Qwen3 model families—each tested at 7B/8B, 14B, and 32B parameter sizes \cite{qwen2, qwen3technicalreport}. To ensure the generality of our findings across architectures and capacities, we included both moderate-sized (7–14B) and larger (27–32B) variants. Appendix~\ref{sec:appendix} reports implementation and decoding configurations.

\subsubsection*{Datasets \& Tasks}
We study \textbf{cross-domain reasoning} between \textbf{mathematical} and \textbf{logical} tasks, evaluating model generalization in both directions. We use \textbf{GSM8K} \cite{cobbe2021gsm8k} for mathematical reasoning, and \textbf{PrOntoQA} \cite{saparov2022language} and \textbf{FOLIO} \cite{han2022folio} for logical reasoning.

\begin{table}[t]
\centering

\resizebox{\columnwidth}{!}{
\begin{tabular}{llccc
}
\toprule
 Model & {DIN} & {Random} & $\Delta$ & Significant  \\
\midrule
LLaMA3.1-8B  & $ \textbf{62.7}_{\pm 2.3}$ & $ 60.3_{\pm 1.8}$ & +2.4 & $\blacktriangledown$  \\
Qwen2.5-7B  & $\textbf{62.8}_{\pm 2.0}$ & $59.5_{\pm 1.6}$ & +3.3 & $\blacktriangledown$ \\
Qwen2.5-14B  & $\textbf{70.7}_{\pm 1.2}$ & $68.8_{\pm 1.2}$ & +1.8 & $\blacktriangledown$ \\
Qwen3-8B & $\textbf{85.5}_{\pm 0.8}$ & $84.0_{\pm 1.9}$ & +1.5 & $\blacktriangledown$ \\
\bottomrule
\end{tabular}
}
\caption{\textbf{Comparison of cross-domain reasoning (GSM8K $\to$ FOLIO) accuracy between DIN-ICL and random neuron selection across different models}. $\Delta$ denotes accuracy gain, and $\blacktriangledown$ indicates statistically significant improvement ($p<0.05$).}
\label{tab:vsdin-matrix}
\end{table}

\begin{table*}[ht!]
\centering
\resizebox{\textwidth}{!}{
\setlength{\tabcolsep}{4pt}
\renewcommand{\arraystretch}{1.1}
\begin{tabular}{llc|ccccc}
\toprule
 \multirow{2}{*}{\textbf{Method}} & \multirow{2}{*}{\textbf{Model Series}} & \multirow{2}{*}{\textbf{Parameters}} & \multicolumn{5}{c}{\textbf{Source Domain $\rightarrow$ Target Domain}} \\
\cmidrule(lr){4-8}
 &  &  & \textbf{FOL$\rightarrow$GSM} & \textbf{GSM$\rightarrow$FOL} & \textbf{GSM$\rightarrow$PRO} & \textbf{PRO$\rightarrow$GSM} & \textbf{Average} \\
\midrule
\multirow{9}{*}{\textsc{Zero-shot}} 
 & \multirow{3}{*}{Qwen-2.5} & 7B   & \best{91.1} & \secondbest{61.8} & 95.6 & \secondbest{89.9} & \secondbest{84.6} \\
 &  & 14B  & \best{93.4} & \secondbest{67.4} & 91.8 & \best{94.3} & 84.2 \\
 &  & 32B  & 91.7 & \secondbest{70.3} & \best{99.8} & \secondbest{91.7} & \secondbest{87.3} \\
\cmidrule(lr){2-8}
 & \multirow{3}{*}{Qwen-3} & 8B     & 93.3 & \secondbest{81.7} & \best{100.0} & 92.4 & \secondbest{91.8} \\
 &  & 14B    & 89.9 & 84.5 & 96.2 & 90.5 & \secondbest{90.2} \\
 &  & 32B    & \secondbest{94.6} & 82.2 & \best{100.0} & \secondbest{94.6} & 92.3 \\
\cmidrule(lr){2-8}
 & \multirow{2}{*}{Gemma-3} & 12B   & \best{93.7} & 61.0 & \best{98.8} & \best{93.2} & \secondbest{84.5} \\
 &  & 27B   & \secondbest{94.3} & 67.9 & 98.2 & \best{94.6} & \secondbest{88.75} \\
\cmidrule(lr){2-8}
 & LLaMA-3.1 & 8B  & \best{81.6} & 56.3 & \best{88.8} & \best{81.7} & \secondbest{77.1} \\
\midrule
\multirow{9}{*}{\textsc{X-ICL}} 
 & \multirow{3}{*}{Qwen-2.5} & 7B   & 89.6$_{\textcolor{blue}{-1.5}}$ & 59.7$_{\textcolor{blue}{-2.1}}$ & 95.4$_{\textcolor{blue}{-0.2}}$ & 89.2$_{\textcolor{blue}{-0.7}}$ & 83.5$_{\textcolor{blue}{-1.1}}$ \\
 &  & 14B  & 92.9$_{\textcolor{blue}{-0.5}}$ & \secondbest{67.4}$_{\textcolor{red}{+0.0}}$ & \secondbest{94.2}$_{\textcolor{red}{+2.4}}$ & 93.2$_{\textcolor{blue}{-1.1}}$ & \secondbest{84.8}$_{\textcolor{red}{+0.6}}$ \\
 &  & 32B  & 91.6$_{\textcolor{blue}{-0.1}}$ & 66.0$_{\textcolor{blue}{-4.3}}$ & 99.6$_{\textcolor{blue}{-0.2}}$ & 91.6$_{\textcolor{blue}{-0.1}}$ & 85.7$_{\textcolor{blue}{-1.6}}$ \\
\cmidrule(lr){2-8}
 & \multirow{3}{*}{Qwen-3} & 8B     & 93.1$_{\textcolor{blue}{-0.2}}$ & 81.2$_{\textcolor{blue}{-0.5}}$ & \secondbest{99.6}$_{\textcolor{blue}{-0.4}}$ & 92.2$_{\textcolor{blue}{-0.2}}$ & 91.5$_{\textcolor{blue}{-0.4}}$ \\
 &  & 14B    & \secondbest{89.9}$_{\textcolor{red}{+0.0}}$ & 83.1$_{\textcolor{blue}{-1.4}}$ & 94.8$_{\textcolor{blue}{-1.4}}$ & \secondbest{91.1}$_{\textcolor{red}{+0.6}}$ & 89.3$_{\textcolor{blue}{-0.9}}$ \\
 &  & 32B    & \secondbest{94.6}$_{\textcolor{red}{+0.0}}$ & \secondbest{83.6}$_{\textcolor{red}{+1.4}}$ & \best{100.0}$_{\textcolor{red}{+0.0}}$ & \secondbest{94.6}$_{\textcolor{red}{+0.0}}$ & \secondbest{92.7}$_{\textcolor{red}{+0.4}}$ \\
\cmidrule(lr){2-8}
 & \multirow{2}{*}{Gemma-3} & 12B   & 92.6$_{\textcolor{blue}{-1.1}}$ & \secondbest{62.5}$_{\textcolor{red}{+1.5}}$ & 97.8$_{\textcolor{blue}{-1.0}}$ & 92.1$_{\textcolor{blue}{-1.1}}$ & 84.3$_{\textcolor{blue}{-0.2}}$ \\
 &  & 27B   & 93.1$_{\textcolor{blue}{-1.2}}$ & \secondbest{68.4}$_{\textcolor{red}{+0.5}}$ & 97.6$_{\textcolor{blue}{-0.6}}$ & 93.1$_{\textcolor{blue}{-1.5}}$ & 88.1$_{\textcolor{blue}{-0.7}}$ \\
\cmidrule(lr){2-8}
 & LLaMA-3.1 & 8B  & 81.4$_{\textcolor{blue}{-0.2}}$ & 55.5$_{\textcolor{blue}{-0.8}}$ & 81.0$_{\textcolor{blue}{-7.8}}$ & 80.4$_{\textcolor{blue}{-1.3}}$ & 74.6$_{\textcolor{blue}{-2.5}}$ \\
\midrule
\multirow{9}{*}{\textsc{DIN-ICL (Ours)}} 
 & \multirow{3}{*}{Qwen-2.5} & 7B   & \secondbest{89.7}$_{\textcolor{blue}{-1.4}}$ & \best{63.5}$_{\textcolor{red}{+1.7}}^\blacktriangledown$ & \best{96.8}$_{\textcolor{red}{+1.2}}^\blacktriangledown$ & \best{90.3}$_{\textcolor{red}{+0.4}}^\blacktriangledown$ & \best{85.1}$_{\textcolor{red}{+0.5}}$ \\
 &  & 14B  & \best{93.4}$_{\textcolor{red}{+0.0}}$ & \best{70.4}$_{\textcolor{red}{+3.0}}^\blacktriangledown$ & \best{94.8}$_{\textcolor{red}{+3.0}}^\blacktriangledown$ & \secondbest{94.0}$_{\textcolor{blue}{-0.3}}$ & \best{86.2}$_{\textcolor{red}{+2.0}}$ \\
 &  & 32B  & \best{92.1}$_{\textcolor{red}{+0.4}}$ & \best{71.4}$_{\textcolor{red}{+1.1}}^\blacktriangledown$ & \secondbest{99.6}$_{\textcolor{blue}{-0.2}}$ & \best{92.1}$_{\textcolor{red}{+0.4}}$ & \best{87.7}$_{\textcolor{red}{+0.4}}$ \\
\cmidrule(lr){2-8}
 & \multirow{3}{*}{Qwen-3} & 8B     & \best{94.6}$_{\textcolor{red}{+1.3}}^\blacktriangledown$ & \best{85.8}$_{\textcolor{red}{+4.1}}^\blacktriangledown$ & \best{100.0}$_{\textcolor{red}{+0.0}}$ & \best{93.0}$_{\textcolor{red}{+0.6}}^\blacktriangledown$ & \best{93.3}$_{\textcolor{red}{+1.5}}$ \\
 &  & 14B    & \best{91.0}$_{\textcolor{red}{+1.1}}^\blacktriangledown$ & \best{85.0}$_{\textcolor{red}{+0.5}}$ & \best{97.0}$_{\textcolor{red}{+0.8}}$ & \best{91.2}$_{\textcolor{red}{+0.7}}^\blacktriangledown$ & \best{91.0}$_{\textcolor{red}{+0.8}}$ \\
 &  & 32B    & \best{95.0}$_{\textcolor{red}{+0.4}}$ & \best{84.0}$_{\textcolor{red}{+1.8}}^\blacktriangledown$ & \best{100.0}$_{\textcolor{red}{+0.0}}$ & \best{95.0}$_{\textcolor{red}{+0.4}}$ & \best{93.0}$_{\textcolor{red}{+0.7}}$ \\
\cmidrule(lr){2-8}
 & \multirow{2}{*}{Gemma-3} & 12B   & \best{93.7}$_{\textcolor{red}{+0.0}}$ & \best{65.5}$_{\textcolor{red}{+4.5}}^\blacktriangledown$ & \best{99.0}$_{\textcolor{red}{+0.2}}$ & \secondbest{92.7}$_{\textcolor{blue}{-0.5}}$ & \best{86.1}$_{\textcolor{red}{+1.6}}$ \\
 &  & 27B   & \best{95.1}$_{\textcolor{red}{+0.8}}^\blacktriangledown$ & \best{68.9}$_{\textcolor{red}{+1.0}}$ & \best{99.2}$_{\textcolor{red}{+1.0}}^\blacktriangledown$ & \secondbest{93.9}$_{\textcolor{blue}{-0.7}}$ & \best{89.3}$_{\textcolor{red}{+0.6}}$ \\
\cmidrule(lr){2-8}
 & LLaMA-3.1 & 8B  & \secondbest{81.5}$_{\textcolor{blue}{-0.1}}$ & \best{63.3}$_{\textcolor{red}{+7.0}}^\blacktriangledown$ & \secondbest{88.6}$_{\textcolor{blue}{-0.2}}$ & \secondbest{81.6}$_{\textcolor{blue}{-0.1}}$ & \best{78.7}$_{\textcolor{red}{+1.6}}$ \\
\bottomrule
\end{tabular}
}
\caption{\textbf{Performance (\%) across cross-domain tasks using different ICL strategies.} 
Each cell under \textsc{DIN-ICL (Ours)} includes a delta compared to the corresponding \textsc{Zero-shot} result. 
The final column reports average accuracy across tasks, where \best{underlined bold} denotes the best and \secondbest{italic} denotes the second best. Significance testing was assessed via an unequal variances t-test in
comparison with \textsc{Zero-Shot}: $\blacktriangledown$ represents a p-value lower than 0.05.}
\label{tab:din_avg}
\end{table*}

\subsubsection*{Baselines}
To benchmark the effectiveness of our proposed framework, we compare against recent and representative methods:

$\bullet$ \textbf{Zero-Shot} \cite{wei2022chain} performs the target task without any in-context examples, relying only on its pretrained knowledge.

$\bullet$ \textbf{X-ICL} \cite{he-etal-2024-using} enhances in-context learning by using LLM-generated natural language explanations to improve model performance.

$\bullet$ \textbf{Set-BSR} \cite{gupta-etal-2023-coverage} greedily selects examples to maximize token-level semantic coverage of the query based on bidirectional similarity.

\subsection{Existence and Importance of DIN (\textbf{RQ1})}
To assess the existence and functional importance of domain-invariant neurons (DIN), we perform pruning analysis on LLaMA-3.1-8B-Instruct as described in Section~\ref{DIN_ident}. For each of the last six layers ($\ell = -6$ to $-1$), we compare the perplexity (PPL) increase from pruning DINs versus random dimensions of equal size, averaging over 300 trials. Statistical significance is evaluated using both empirical $p$-values and normal approximation.

Results show that pruning DINs consistently leads to greater degradation than random pruning. In the source domain, DIN pruning causes 5.2–8.1 $\%$ relative increase in PPL across layers ($\ell=-6$ to $-2$), significantly exceeding the random pruning baseline (at $\ell=-6$, PPL rises by +7.99$\%$ when pruning DINs, compared with only +0.07$\%$ under random pruning ($p_{\text{emp}}=0.0332$)). Consistently, Table~\ref{tab:vsdin-matrix} shows that using DIN-selected neurons for cross-domain in-context learning yields statistically significant accuracy gains over random neuron selection across multiple models.

Overall, pruning DINs from layers $-6$ to $-1$ leads to significantly greater degradation than random pruning in both domains, confirming that a compact set of domain-invariant neurons are both identifiable and functionally important for cross-domain generalization.

\begin{table*}[ht!]
\centering
\resizebox{\textwidth}{!}{
\setlength{\tabcolsep}{4pt}
\renewcommand{\arraystretch}{1.1}
\begin{tabular}{llc|ccccc}
\toprule
 \multirow{2}{*}{\textbf{Method}} & \multirow{2}{*}{\textbf{Model Series}} & \multirow{2}{*}{\textbf{Parameters}} & \multicolumn{5}{c}{\textbf{Source Domain $\rightarrow$ Target Domain}} \\
\cmidrule(lr){4-8}
 &  &  & \textbf{FOL$\rightarrow$GSM} & \textbf{GSM$\rightarrow$FOL} & \textbf{GSM$\rightarrow$PRO} & \textbf{PRO$\rightarrow$GSM} & \textbf{Average} \\
\midrule
\multirow{9}{*}{\textsc{Set-BSR}} 
 & \multirow{3}{*}{Qwen-2.5} & 7B   & \secondbest{89.6} & \secondbest{59.7} & \secondbest{95.4} & \secondbest{89.2} & \secondbest{83.5} \\
 &  & 14B  & \secondbest{92.9} & \secondbest{67.4} & \best{94.2} & \secondbest{93.2} & \secondbest{84.8} \\
 &  & 32B  & \secondbest{91.6} & \secondbest{66.0} & \best{99.6} & \secondbest{91.6} & \secondbest{85.7} \\
\cmidrule(lr){2-8}
 & \multirow{3}{*}{Qwen-3} & 8B     & \secondbest{93.1} & \secondbest{81.2} & \secondbest{99.6} & \secondbest{92.2} & \secondbest{91.5} \\
 &  & 14B    & \secondbest{89.9} & \secondbest{83.1} & \secondbest{94.8} & \best{91.1} & \secondbest{89.3} \\
 &  & 32B    & \secondbest{94.6} & \secondbest{83.6} & \best{100.0} & \secondbest{94.6} & \secondbest{92.7} \\
\cmidrule(lr){2-8}
 & \multirow{2}{*}{Gemma-3} & 12B   & \secondbest{92.6} & \secondbest{62.5} & \secondbest{97.8} & \secondbest{92.1} & \secondbest{84.3} \\
 &  & 27B   & \secondbest{93.1} & \secondbest{68.4} & \secondbest{97.6} & \best{93.1} & \secondbest{88.1} \\
\cmidrule(lr){2-8}
 & LLaMA-3.1 & 8B  & \secondbest{81.4} & \secondbest{55.5} & \secondbest{81.0} & \secondbest{80.4} & \secondbest{74.6} \\
\midrule
\multirow{9}{*}{\textsc{DIN-ICL (Ours)}} 
 & \multirow{3}{*}{Qwen-2.5} & 7B   & \best{89.7}$_{\textcolor{red}{+0.1}}$ & \best{63.5}$_{\textcolor{red}{+3.8}}^\blacktriangledown$ & \best{96.8}$_{\textcolor{red}{+1.4}}^\blacktriangledown$ & \best{90.3}$_{\textcolor{red}{+1.1}}^\blacktriangledown$ & \best{85.1}$_{\textcolor{red}{+1.6}}^\blacktriangledown$ \\
 &  & 14B  & \best{93.4}$_{\textcolor{red}{+0.5}}$ & \best{70.4}$_{\textcolor{red}{+3.0}}^\blacktriangledown$ & \secondbest{93.9}$_{\textcolor{blue}{-0.3}}$ & \best{94.0}$_{\textcolor{red}{+0.8}}$ & \best{86.2}$_{\textcolor{red}{+1.1}}^\blacktriangledown$ \\
 &  & 32B  & \best{92.1}$_{\textcolor{red}{+0.5}}$ & \best{71.4}$_{\textcolor{red}{+5.4}}^\blacktriangledown$ & \best{99.6}$_{\textcolor{red}{+0.0}}$ & \best{92.1}$_{\textcolor{red}{+0.5}}$ & \best{87.7}$_{\textcolor{red}{+2.0}}^\blacktriangledown$ \\
\cmidrule(lr){2-8}
 & \multirow{3}{*}{Qwen-3} & 8B     & \best{94.6}$_{\textcolor{red}{+1.5}}^\blacktriangledown$ & \best{85.8}$_{\textcolor{red}{+4.6}}^\blacktriangledown$ & \best{100.0}$_{\textcolor{red}{+0.4}}$ & \best{93.0}$_{\textcolor{red}{+0.8}}$ & \best{93.3}$_{\textcolor{red}{+1.8}}^\blacktriangledown$ \\
 &  & 14B    & \best{91.0}$_{\textcolor{red}{+1.1}}^\blacktriangledown$ & \best{85.0}$_{\textcolor{red}{+1.9}}^\blacktriangledown$ & \best{97.0}$_{\textcolor{red}{+2.2}}^\blacktriangledown$ & \secondbest{90.9}$_{\textcolor{blue}{-0.2}}$ & \best{90.8}$_{\textcolor{red}{+1.5}}^\blacktriangledown$ \\
 &  & 32B    & \best{95.0}$_{\textcolor{red}{+0.4}}$ & \best{84.0}$_{\textcolor{red}{+0.4}}$ & \best{100.0}$_{\textcolor{red}{+0.0}}$ & \best{95.0}$_{\textcolor{red}{+0.4}}$ & \best{93.0}$_{\textcolor{red}{+0.3}}$ \\
\cmidrule(lr){2-8}
 & \multirow{2}{*}{Gemma-3} & 12B   & \best{93.7}$_{\textcolor{red}{+1.1}}^\blacktriangledown$ & \best{65.5}$_{\textcolor{red}{+3.0}}^\blacktriangledown$ & \best{99.0}$_{\textcolor{red}{+1.2}}^\blacktriangledown$ & \best{92.7}$_{\textcolor{red}{+0.6}}$ & \best{86.1}$_{\textcolor{red}{+1.8}}^\blacktriangledown$ \\
 &  & 27B   & \best{95.1}$_{\textcolor{red}{+2.0}}^\blacktriangledown$ & \best{68.9}$_{\textcolor{red}{+0.5}}$ & \best{99.2}$_{\textcolor{red}{+1.6}}^\blacktriangledown$ & \secondbest{92.8}$_{\textcolor{blue}{-0.3}}$ & \best{89.0}$_{\textcolor{red}{+0.9}}^\blacktriangledown$ \\
\cmidrule(lr){2-8}
 & LLaMA-3.1 & 8B  & \best{81.5}$_{\textcolor{red}{+0.1}}$ & \best{63.3}$_{\textcolor{red}{+7.8}}^\blacktriangledown$ & \best{88.6}$_{\textcolor{red}{+7.6}}^\blacktriangledown$ & \best{81.6}$_{\textcolor{red}{+1.2}}^\blacktriangledown$ & \best{78.7}$_{\textcolor{red}{+4.1}}^\blacktriangledown$ \\
\bottomrule
\end{tabular}
}
\caption{\textbf{Comparison between retrieval-based ICL (Set-BSR) and DIN-ICL.}}
\label{tab:setbsr_est}
\end{table*}

\begin{figure}[t!]
  \includegraphics[scale=0.23]{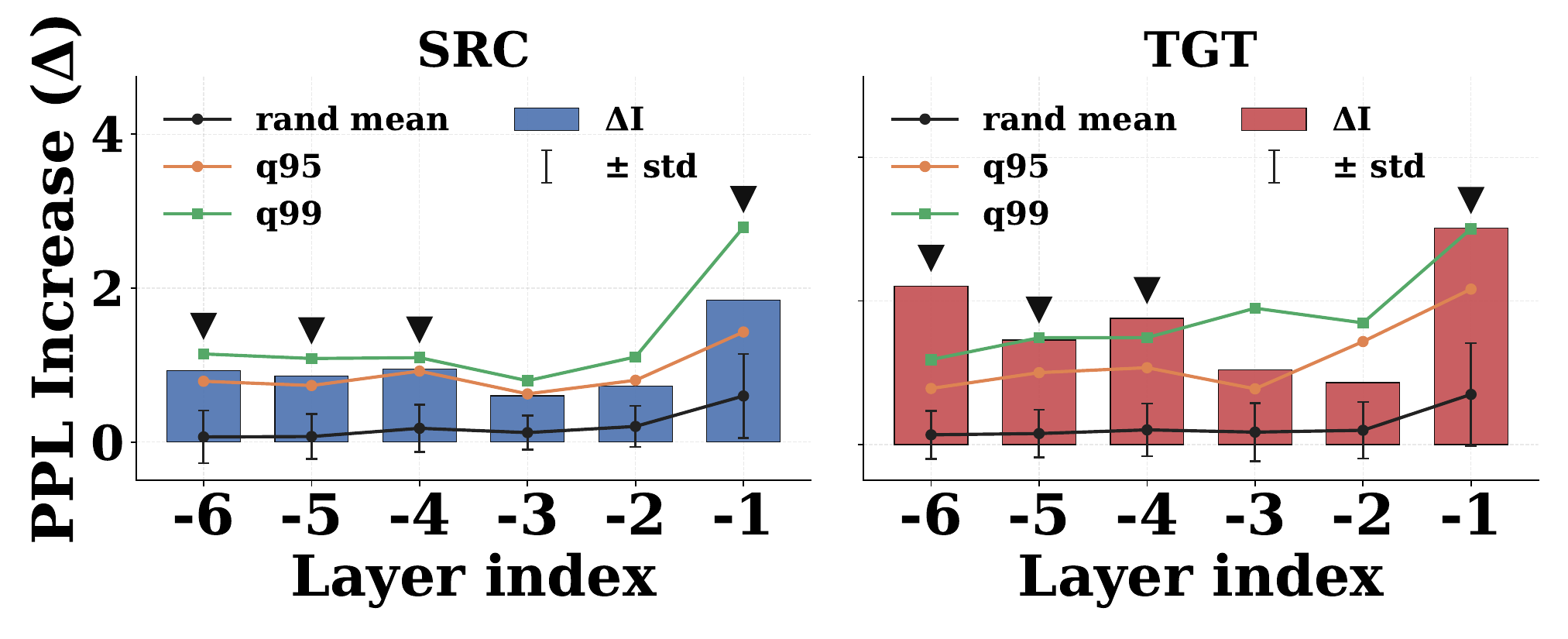}
  \caption{\textbf{Perplexity increase from pruning DINs vs. random neurons across the last six layers.} Results are averaged over 300 trials. The solid line denotes mean PPL increase after pruning DINs, while the dashed line and shaded areas indicate the random pruning baseline (mean with 95th and 99th percentiles) $\blacktriangledown$ indicates statistically significant improvement ($p<0.05$).}
  \label{lesion}
\end{figure}


\subsection{Cross-domain ICl improvement (\textbf{RQ2})}

To evaluate whether DIN can support improved generalization in cross-domain reasoning, we conduct a large-scale comparative study across diverse model families, sizes, and transfer directions. Specifically, we compare three ICL strategies: \textsc{Zero-shot} CoT prompting without in-context examples, \textsc{X-ICL} strengthens ICL by using LLM-generated explanations to build more robust demonstration prompts, and our proposed \textsc{DIN}-guided retrieval, which operates within subspaces defined by consistent cross-domain activation patterns.

Table~\ref{tab:din_avg} reports accuracy results on four transfer directions: FOLIO$\rightarrow$GSM8K, GSM8K$\rightarrow$FOLIO, GSM8K$\rightarrow$PrOntoQA, and PrOntoQA$\rightarrow$GSM8K, across nine open-source models ranging from 7B to 32B parameters. In all cases, DIN-based retrieval either matches or outperforms the baselines, with especially pronounced gains in more challenging settings such as GSM$\rightarrow$FOL (e.g., +3.0 on Qwen2.5-14B and +4.1 on Qwen3-8B) and Pronto$\rightarrow$GSM (e.g., +0.6 on Qwen3-8B and +0.4 on Qwen2.5-7B). Compared to \textsc{Zero-shot}, \textsc{DIN} achieves an average gain of +0.5–2.0 points on most Qwen and Gemma models, and notably outperforms \textsc{X-ICL} across all sizes of LLaMA3.1, where the latter shows degradation (e.g., -7.8 on GSM$\rightarrow$PRO) while DIN remains stable.

These improvements suggest that DIN identifies a more transferable latent subspace for selecting effective demonstrations in ICL, especially when domain shift is significant. Notably, models with larger capacity (e.g., Qwen3-32B) show strong baseline performance where DIN yields smaller gains (e.g., +0.4 or no change), whereas smaller or less robust models benefit more from DIN-guided selection. This pattern further supports the hypothesis that DIN acts as a stabilizing inductive bias under domain mismatch.

Table \ref{tab:setbsr_est} compares our proposed DIN-ICL with the non-parametric Set-BSR baseline \cite{gupta-etal-2023-coverage} across four transfer directions (FOL→GSM, GSM→FOL, GSM→Pro, and Pro→GSM) and multiple model families. Overall, DIN-ICL consistently outperforms Set-BSR across all model scales and domains, confirming the effectiveness of identifying domain-invariant neurons for cross-domain reasoning. While Set-BSR already improves over standard token-level retrieval by maximizing coverage of query semantics, it remains sensitive to domain-specific embedding shifts.
By contrast, DIN-ICL aligns retrieval within a neuron-stable subspace, effectively reducing variance in cross-domain similarity estimation. On average, DIN-ICL achieves +0.8–1.5 pp higher accuracy than Set-BSR and up to +3 pp gains in high-shift settings such as GSM→FOL.
Notably, the performance gap diminishes as model size grows, suggesting that large LMs already encode partially domain-invariant representations, yet DIN-ICL further stabilizes retrieval, yielding the most consistent improvements across all configurations.

\begin{figure}[t!]
  \includegraphics[scale=0.15]{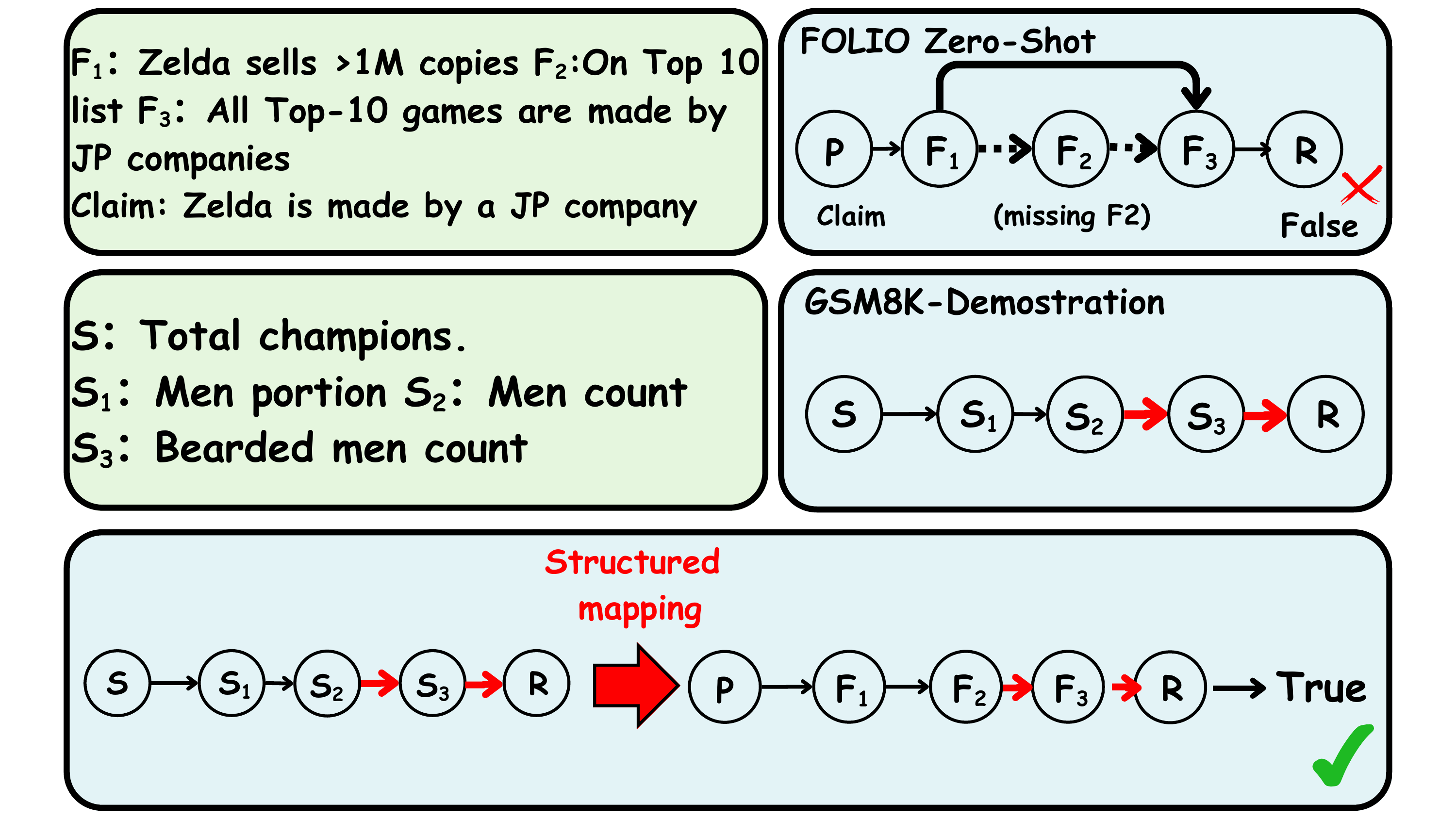}
  \caption{Case study illustrating a linear reasoning topology.}
  \label{case1}
\end{figure}

In summary, DIN-based retrieval provides consistent, model-agnostic improvements in cross-domain in-context reasoning and outperforms traditional full-space similarity approaches, particularly when zero-shot performance is weak or unstable.

\subsection{Structural Case Analysis of Cross-Domain Reasoning Failures (\textbf{RQ3})}

\begin{figure}[t!]
  \includegraphics[scale=0.15]{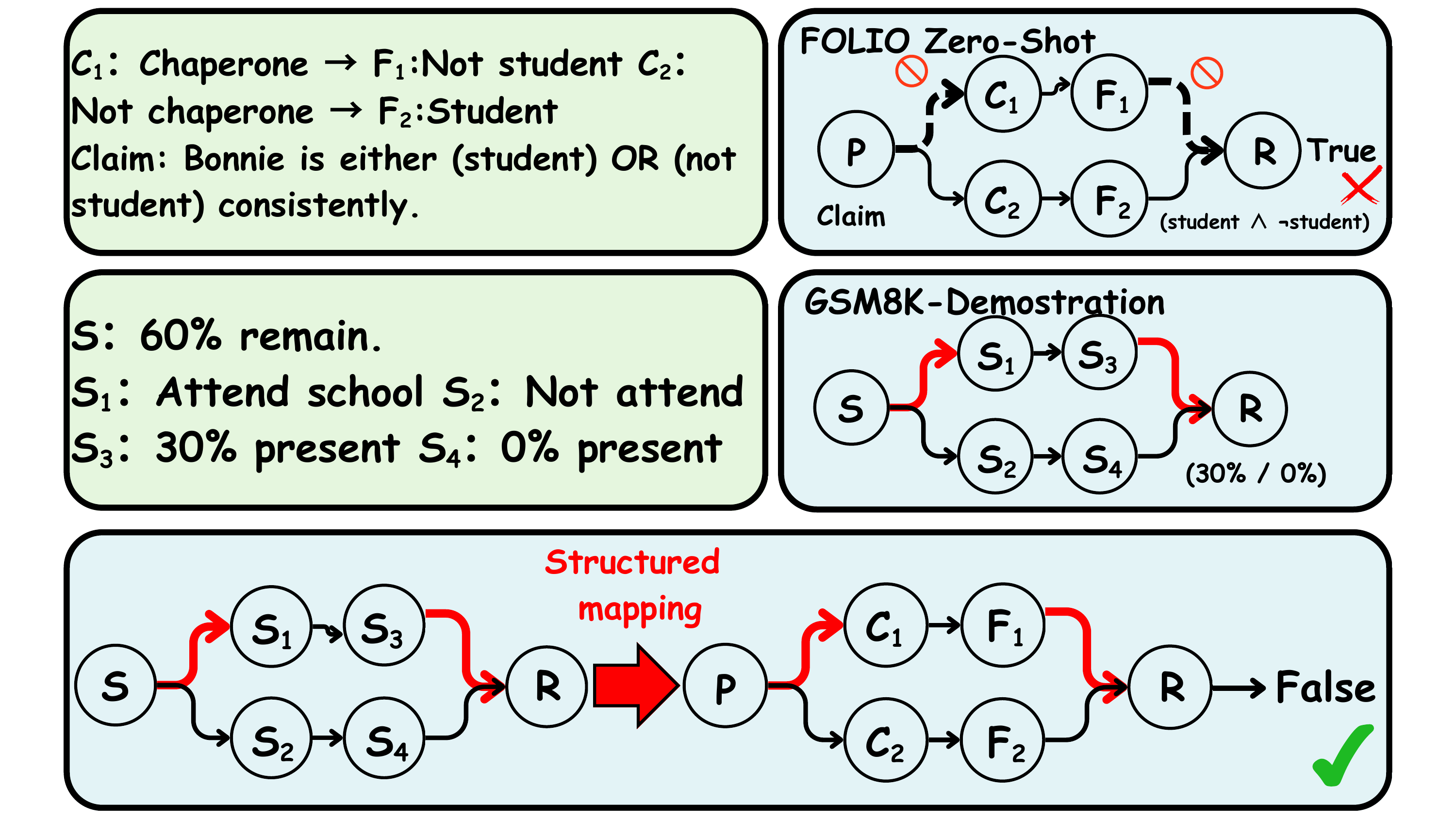}
  \caption{Case study illustrating a binary branching reasoning topology}
  \label{case2}
\end{figure}

\begin{figure}[t!]
  \includegraphics[scale=0.15]{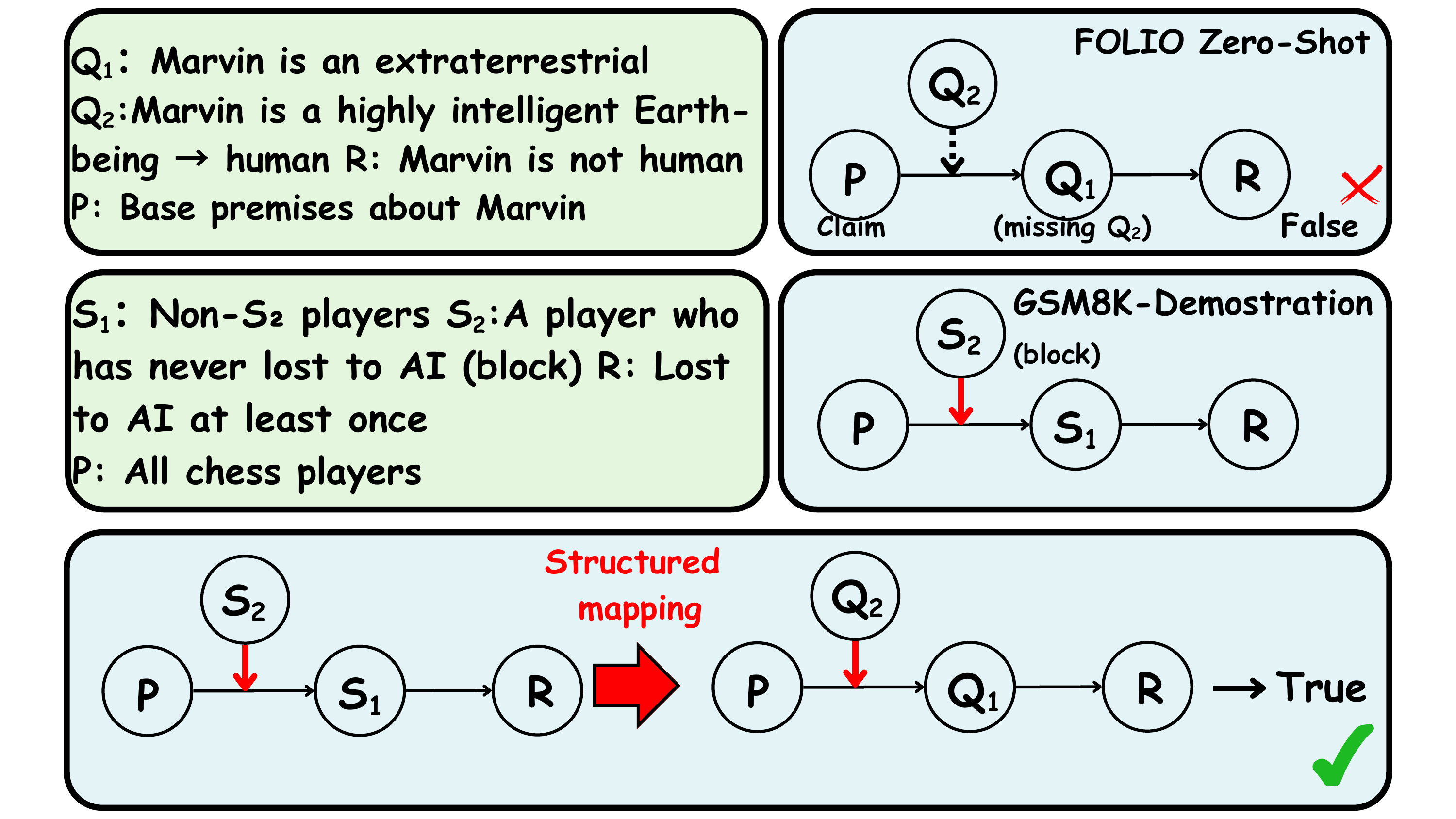}
  \caption{Case study illustrating a blocking-condition reasoning topology.}
  \label{case3}
\end{figure}

Figures~\ref{case1}, \ref{case2}, and \ref{case3} present representative examples illustrating how DIN-ICL repairs broken reasoning topologies under cross-domain transfer.

In the first case, the zero-shot model fails to perform correct case analysis, as it collapses two conflicting branches into a single path, thereby missing the required multi-branch structure. The demonstration retrieved via DIN-retrieval provides an isomorphic two-branch topology, enabling structured mapping and allowing the model to recover the correct branch separation.

The second case shows a chained reasoning failure, where the zero-shot model omits a necessary intermediate implication. The linear step-by-step structure in the GSM8K demo supplies an appropriate scaffold, helping the model reconstruct the missing link. Finally, Figure~\ref{case3} illustrates a blocking-condition topology. The zero-shot model ignores the blocking branch, leading to an invalid conclusion. DIN-ICL retrieves a GSM8K demonstration with the same blocking structure, enabling the model to reinstate the blocked path and reach the correct inference. All complete examples can be found in the table in the Appendix \ref{example}.

Together, these cases demonstrate that DIN-ICL boosts cross-domain reasoning by supplying demonstrations whose internal structures match the required reasoning topology of the target query.

\section{Related Work}

\subsection{Generalization in LLM}
Large language models (LLMs) often degrade under domain shifts \cite{oncel2024adaptation, oh2025understanding}. Existing approaches—such as data-centric adaptation \cite{wang2024neurons}, prompt calibration \cite{zhao2021calibrate, honda-oka-2025-exploring, he-etal-2024-using}, and parameter-efficient tuning \cite{hu2022lora}—primarily modify data or prompts, while overlooking the model’s internal transferability. Recent work has begun examining cross-domain representation alignment \cite{aghajanyan2020intrinsic}, including neuron-level alignment in multilingual settings \cite{huang2025neurons}. Although neuron-level analyses are well explored \cite{chen2024analyzing, sajjad2022neuron}, the existence and role of domain-invariant neurons under domain shift remain unknown. We address this gap by leveraging domain-invariant neurons (DINs) to improve cross-domain generalization.

\subsection{Neuron-Level Analysis and Functional Attribution}
Understanding the functional roles of individual neurons has been central to interpretability research \cite{chen2025identifying, sajjad2022neuron, antverg2021pitfalls}. Prior work has identified knowledge-related neurons \cite{dai2021knowledge} and memory behaviour in feed-forward layers \cite{geva2020transformer}, while lesion-based methods quantify the contribution of specific components \cite{voita2019bottom, meng2022locating, li2024optimal} or confidence-regulation neurons \cite{stolfo2024confidence}.
In contrast, we identify domain-invariant neurons using cross-domain z-score polarity consistency and show via lesion tests that removing them substantially degrades performance, highlighting their importance for cross-domain reasoning.

\subsection{Example Selection for In-Context Learning}
In-context learning (ICL) is highly sensitive to demonstration selection \cite{luo2024context}. Existing retrieval methods rely on semantic similarity \cite{rubin2021learning}, dense retrievers \cite{wang2023learning}, uncertainty signals \cite{ling2024uncertainty, huang2024unlocking, margatina2023active}, coverage-based selection \cite{gupta-etal-2023-coverage}, or MMR-based diversification \cite{liu2023lost}, but largely operate on surface-level or input-level cues. Recent work has begun using internal token representations \cite{liu2023lost}. In contrast, our method retrieves demonstrations in the DIN vector space, leveraging domain-robust internal dimensions for more stable and transferable prompting.

\section{Conclusion}
We presented DIN-ICL, a framework that leverages Domain-Invariant Neurons (DINs) to improve cross-domain in-context learning. By identifying neurons with consistent activation polarity across domains and using them to form a DIN-based retrieval subspace, our method selects demonstrations that capture transferable reasoning structure. Experiments across multiple models and math–logic transfer settings show that DIN-ICL consistently improves cross-domain accuracy over zero-shot and strong retrieval baselines while maintaining in-domain performance. These results highlight neuron-level invariance as a useful inductive bias for robust cross-domain reasoning.

\section*{Limitations}
First, our DIN identification method uses a simple polarity-consistency rule and fixed thresholds, which may not capture more complex invariance. Second, experiments are limited to reasoning domains (GSM8K, PrOntoQA, FOLIO); broader domains should be explored. The causal role of identified neurons remains preliminary, and observed gains, though consistent, are modest. Future work may integrate DIN-guided retrieval with adaptive or fine-tuning-based methods for stronger cross-domain generalization.


\bibliography{custom}

\clearpage
\appendix

\section{Appendix}
\label{sec:appendix}

\subsection{Implementation Details}
\subsubsection{Model Inference}
All experiments are conducted using vLLM \cite{kwon2023efficient} as the inference backend to ensure efficient serving of large models and fast hidden-state extraction. Unless otherwise specified, model precision is set to FP16, following the default mixed-precision configuration of vLLM. We use HuggingFace Transformers \cite{wolf-etal-2020-transformers} for model loading, tokenization, and hidden-state access.

\subsubsection{Generation Hyperparameters}
Across all experiments—including cross-domain ICL evaluation, DIN retrieval, and case studies—we use the following decoding configuration:
\begin{table}[h]
\centering

\begin{tabular}{ll}
\toprule
\textbf{Category} & \textbf{Setting} \\
\midrule
Temperature           & \textbf{0.6} \\
Sampling (Top-p/k)   & Top-p = 1.0,\; Top-k = 50 \\
Max Gen Length        & 8192 tokens \\
Random Seed           & 1-30 \\
\bottomrule
\end{tabular}
\caption{Decoding setup used throughout all experiments.}
\end{table}

\subsection{Hyperparameter Analysis}
\begin{figure}[!ht]
  \includegraphics[scale=0.23]{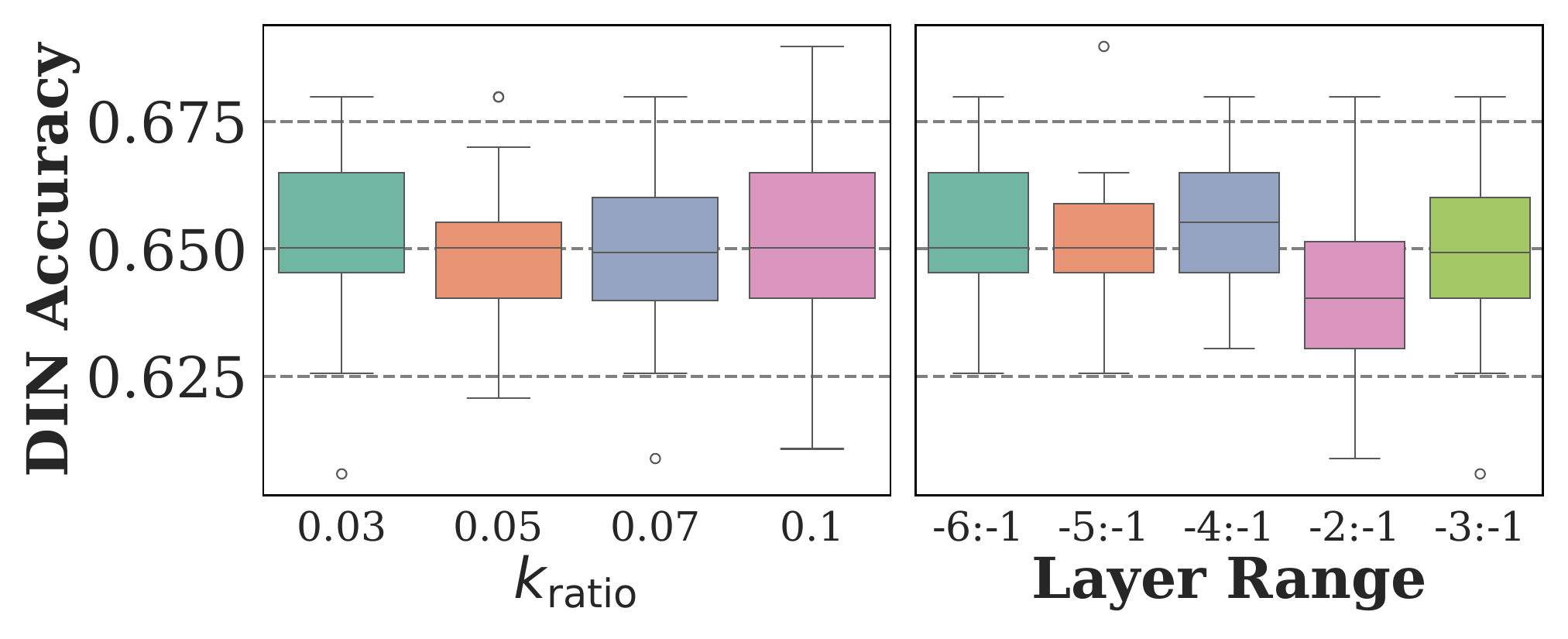}
  \caption{\textbf{Effect of key hyperparameters on DIN-based ICL performance.} Left: Increasing $k_{\mathrm{ratio}}$ generally leads to slightly higher DIN accuracy. Right: DIN subspaces extracted from deeper layers tend to outperform shallower ones.}
  \label{hyper_1}
\end{figure}

We investigate how key hyperparameters affect the effectiveness of DIN selection and its downstream impact on in-context reasoning. Specifically, we analyze the influence of the selection ratio $k_{\mathrm{ratio}}$ and the choice of layer range used to extract domain-invariant neurons.

Figure \ref{hyper_1} (left) shows that increasing $k_{\mathrm{ratio}}$ from 0.03 to 0.1 leads to slightly higher DIN accuracy on average, although the variance remains large. This suggests that using more neurons provides richer signal for cross-domain generalization, but over-selection may introduce noise. Figure \ref{hyper_1} (right) compares different layer ranges, showing that deeper layers consistently yield higher accuracy than shallower ones. This is consistent with prior findings that later layers in LLMs encode more task-specific and transferable representations.

\begin{figure}[t!]
  \includegraphics[scale=0.31]{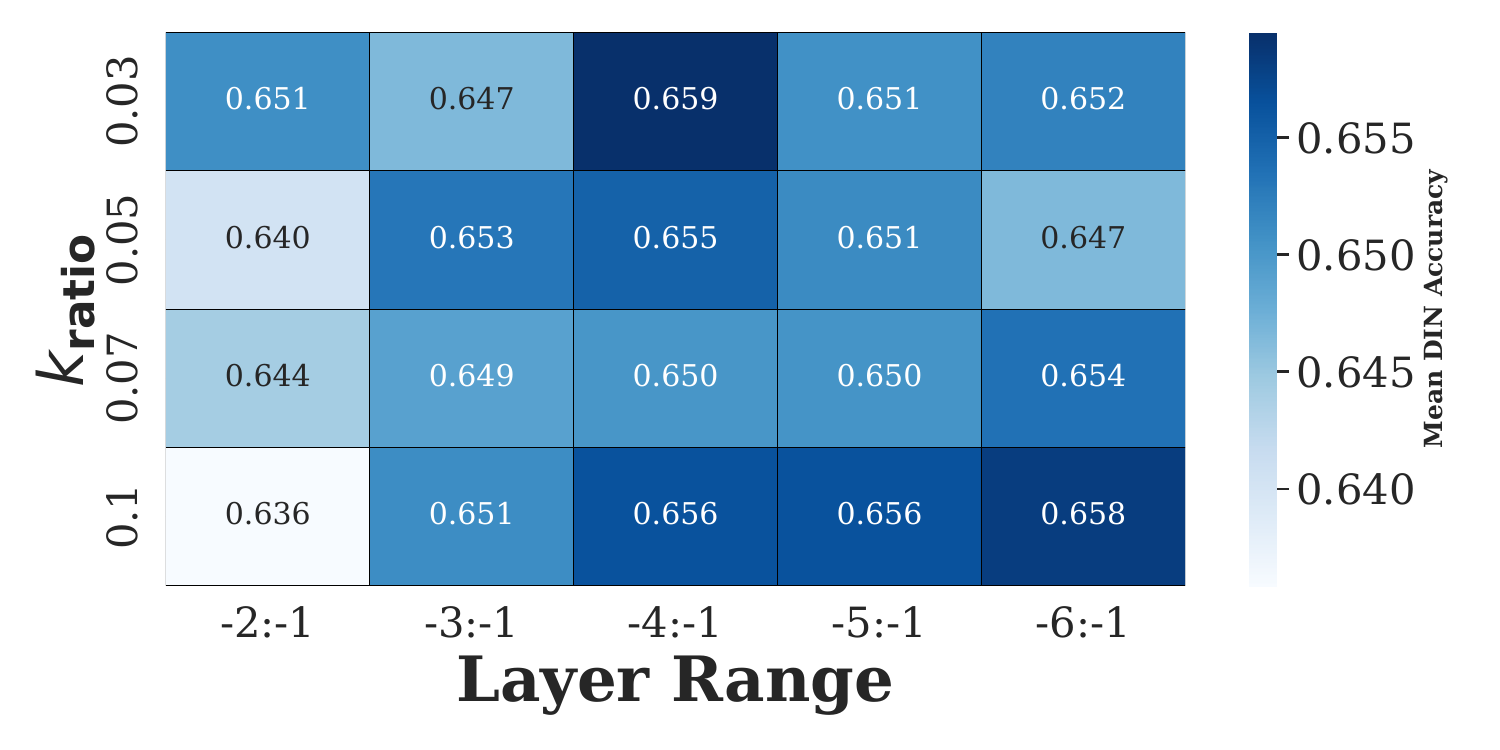}
  \caption{\textbf{Mean DIN accuracy across different combinations of $k_{\mathrm{ratio}}$ and layer range.}}
  \label{hyper_2}
\end{figure}

To jointly analyze the interaction between the two hyperparameters, we plot a heatmap in Figure \ref{hyper_2}. The results confirm that deeper layer ranges and moderate $k_{\mathrm{ratio}}$ values yield the most reliable DIN subspaces across tasks. Notably, the highest DIN accuracy ($0.659$) is achieved with $k_{\mathrm{ratio}}=0.03$ and layer range L$-4$:$-1$, indicating that quality can sometimes outweigh quantity when selecting stable neurons.

These results highlight the importance of careful hyperparameter tuning when applying DIN-based retrieval in practice. We adopt the best-performing settings in subsequent experiments unless otherwise noted.

\subsection{Prompts}
We adopt task-specific system prompts for each dataset to ensure consistent reasoning style and unified answer formatting across domains. All prompts follow a two-stage structure: (1) the model is instructed to provide a short step-by-step reasoning; and (2) the final answer must be output on a separate line using a strict standardized format. This design avoids ambiguity in option extraction and enables reliable automatic evaluation.

\paragraph{PrOntoQA.}
For binary logical reasoning tasks in PrOntoQA, the system prompt is:
\begin{tcolorbox}[
    colback=gray!4,
    colframe=gray!40,
    boxrule=0.3pt,
    arc=2pt,
    left=4pt,right=4pt,top=4pt,bottom=4pt]
\small\texttt{%
You are a careful reasoner. Think step by step concisely. \\
Then on a new line, output exactly: `Final answer: A' or `Final answer: B'.
}
\end{tcolorbox}

\paragraph{FOLIO.}
FOLIO requires three-way classification (entailment / contradiction / unknown).  
We extend the same reasoning template to support three options:
\begin{tcolorbox}[
    colback=gray!4,
    colframe=gray!40,
    boxrule=0.3pt,
    arc=2pt,
    left=4pt,right=4pt,top=4pt,bottom=4pt]
\small\texttt{%
You are a careful reasoner. Think step by step concisely. \\
Then on a new line, output exactly: \\
`Final answer: A' or `Final answer: B' or `Final answer: C'.
}
\end{tcolorbox}

\paragraph{GSM8K.}
For math word problems, we follow the conventional Chain-of-Thought (CoT) prompting format:
\begin{tcolorbox}[
    colback=gray!4,
    colframe=gray!40,
    boxrule=0.3pt,
    arc=2pt,
    left=4pt,right=4pt,top=4pt,bottom=4pt]
\small\texttt{%
You are a careful math reasoner. Solve step by step concisely. \\
Then on a new line, output exactly: `Final answer: <number>'.
}
\end{tcolorbox}

\subsection{Dataset Statistics}
To evaluate cross-domain reasoning, we use three representative datasets covering mathematical and logical tasks: GSM8K, FOLIO, and PrOntoQA. These datasets differ in reasoning style, language form, and label space, offering a diverse benchmark for assessing the transferability of domain-invariant neurons. Table~\ref{tab:dataset-stats} summarizes their key statistics.

\begin{table}[h]
\centering
\begin{tabular}{lccc}
\toprule
\textbf{Dataset} & \textbf{\#Train} & \textbf{\#Test} & \textbf{Labels} \\
\midrule
GSM8K           & 7,473  & 1,319   & Numeric \\
FOLIO         & 5,216  & 500    & A/B/C \\
PrOntoQA  & 22,000 & 2,000 & A/B (Yes/No) \\
\bottomrule
\end{tabular}
\caption{Statistics of datasets used in our cross-domain experiments.}
\label{tab:dataset-stats}
\end{table}

\subsection{Data Example}
\label{example}
\begin{table*}[ht!]
\centering
\small
\begin{tabular}{p{2cm}p{10.5cm}}
\toprule
\textbf{Setting} & \textbf{Model Reasoning and Output} \\
\midrule
\multirow{5}{*}{\textbf{Context}} & A Japanese game company created the game The Legend of Zelda.All games on the Top 10 list are made by Japanese game companies.If a game sells more than one million copies, then it will be included in the Top 10 list.The Legend of Zelda sold more than one million copies. \\
\midrule
\multirow{7}{*}{\textbf{Zero-shot}} &
\textit{Since the Legend of Zelda sold more than one million copies, it should be included in the Top 10 list. However, it is stated that the Legend of Zelda is not on the Top 10 list. This creates a contradiction with the information that all games on the Top 10 list are made by Japanese game companies and the Legend of Zelda is made by a Japanese game company. Therefore, the statement "The Legend of Zelda is not on the Top 10 list" must be true. Final answer: A} \\
& \textbf{Final answer: A (True)} \\
\midrule
\multirow{6}{*}{\textbf{Demostration}} & Question: In the last 25 years, 60$\%$ of Scrabble champions have been women and the rest have been men. There is one champion per year. Of the men, 40$\%$ had a beard. How many men with beards have been Scrabble champion? Reasoning: 40$\%$ of champions are men because 100 - 60 = <<100-60=40>>40 10 champions have been men because 25 x .4 = <<25*.4=10>>10 4 champions have been men with beards because 10 x .4 = <<10*.4=4>>4 \\
\midrule
\multirow{5}{*}{\textbf{DIN-ICL}} &
\textit{Since the Legend of Zelda sold more than one million copies and all games on the Top 10 list are made by Japanese game companies, and the Legend of Zelda is made by a Japanese game company, it must be on the Top 10 list. Therefore, the statement "The Legend of Zelda is not on the Top 10 list" is false. Final answer: B} \\
& \textbf{Final answer: B (False)} \\
\bottomrule
\end{tabular}
\caption{Comparison between zero-shot reasoning and DIN-ICL reasoning on a cross-domain example. DIN-guided retrieval corrects the logical inconsistency made by zero-shot prompting.}
\label{tab:case1}
\end{table*}

\begin{table*}[ht!]
\centering
\small
\begin{tabular}{p{2cm}p{10.5cm}}
\toprule
\textbf{Setting} & \textbf{Model Reasoning and Output} \\
\midrule
\multirow{11}{*}{\textbf{Context}} & People in this club who perform in school talent shows often attend and are very engaged with school events.People in this club either perform in school talent shows often or are inactive and disinterested community members.People in this club who chaperone high school dances are not students who attend the school.All people in this club who are inactive and disinterested members of their community chaperone high school dances.All young children and teenagers in this club who wish to further their academic careers and educational opportunities are students who attend the school. Bonnie is in this club and she either both attends and is very engaged with school events and is a student who attends the school or is not someone who both attends and is very engaged with school events and is not a student who attends the school. \\
\midrule
\multirow{26}{*}{\textbf{Zero-shot}} &
\textit{Let's break down the information step by step:1. If Bonnie chaperones high school dances, then she is not a student who attends the school (from the third statement).2. If Bonnie does not chaperone high school dances, then she performs in school talent shows often (from the second statement).3. If Bonnie performs in school talent shows often, then she is very engaged with school events (from the first statement).4. If Bonnie is very engaged with school events, then she is a student who attends the school (from the first statement).5. From steps 1 and 4, we can conclude that Bonnie cannot be both a student who attends the school and chaperone high school dances.6. From steps 2 and 3, we can conclude that if Bonnie does not chaperone high school dances, then she is a student who attends the school.7. From steps 5 and 6, we can conclude that Bonnie is not a student who attends the school if she chaperones high school dances, and she is a student who attends the school if she does not chaperone high school dances.8. From the fourth statement, we can conclude that if Bonnie is a student who attends the school, then she is a young child or teenager who wishes to further her academic career and educational opportunities.9. From steps 7 and 8, we can conclude that Bonnie is a young child or teenager who wishes to further her academic career and educational opportunities if she does not chaperone high school dances.10. From the fifth statement, we can conclude that if Bonnie is an inactive and disinterested member of the community, then she chaperones high school dances.11. From steps 7 and 10, we can conclude that Bonnie is an inactive and disinterested member of the community if she chaperones high school dances.12. From steps 9 and 11, we can conclude that Bonnie is both a young child or teenager who wishes to further her academic career and educational opportunities and an inactive and disinterested member of the community if she does not chaperone high school dances.Therefore, the statement is true.Final answer: A} \\
& \textbf{Final answer: A (True)} \\
\midrule
\multirow{5}{*}{\textbf{Demostration}} & Question: Forty percent of the students have elected to learn from home during the pandemic. The remaining students are divided into two equal groups, only one of which is physically in school on any day. What percent of students are present in school? Reasoning: 100-40=<<100-40=60>>60 percent of students are remaining. 60/2=<<60/2=30>>30 percent are present in school. \\
\midrule
\multirow{12}{*}{\textbf{DIN-ICL}} &
\textit{Let\'s break down the information step by step:1. If Bonnie chaperones high school dances, then she is not a student who attends the school (from the context).2. If Bonnie does not chaperone high school dances, then she performs in school talent shows often (from the given statement).3. If Bonnie performs in school talent shows often, then she is very engaged with school events (from the context).4. If Bonnie is very engaged with school events, then she is a student who attends the school (from the context).5. From steps 1 and 4, we have a contradiction: Bonnie cannot both be a student who attends the school and not be a student who attends the school.6. Therefore, the given statement "If Bonnie either chaperones high school dances or, if she does not, she performs in school talent shows often" must be false, because it leads to a contradiction.Final answer: B} \\
& \textbf{Final answer: B (False)} \\
\bottomrule
\end{tabular}
\caption{Comparison between zero-shot reasoning and DIN-ICL reasoning on a cross-domain example. DIN-guided retrieval corrects the logical inconsistency made by zero-shot prompting.}
\label{tab:case2}
\end{table*}

\end{document}